\documentclass[conference]{IEEEtran}
\IEEEoverridecommandlockouts
\usepackage{cite}
\usepackage{amsmath,amssymb,amsfonts}
\usepackage{algorithmic}
\usepackage{graphicx}
\usepackage{textcomp}
\usepackage{longtable}
\usepackage[table]{xcolor}
\usepackage{hyperref}
\def\BibTeX{{\rm B\kern-.05em{\sc i\kern-.025em b}\kern-.08em
    T\kern-.1667em\lower.7ex\hbox{E}\kern-.125em}}
\begin{document}

\title{Systematic Evaluation of Online Speaker Diarization Systems Regarding their Latency\\}

\author{\IEEEauthorblockN{1\textsuperscript{st} Roman Aperdannier}
\IEEEauthorblockA{\textit{Faculty of Business} \\
\textit{University of Applied Science}\\
Ansbach, Germany \\
aperdannier19472@hs-ansbach.de}
\and
\IEEEauthorblockN{2\textsuperscript{nd} Sigurd Schacht}
\IEEEauthorblockA{\textit{Faculty of Business} \\
\textit{University of Applied Science}\\
Ansbach, Germany \\
sigurd.schacht@hs-ansbach.de}
\and
\IEEEauthorblockN{3\textsuperscript{rd} Alexander Piazza}
\IEEEauthorblockA{\textit{Faculty of Business} \\
\textit{University of Applied Science}\\
Ansbach, Germany \\
alexander.piazza@hs-ansbach.de}

}

\maketitle

\begin{abstract}
In this paper, different online speaker diarization systems are evaluated on the same hardware with the same test data with regard to their latency. The latency is the time span from audio input to the output of the corresponding speaker label.  As part of the evaluation, various model combinations within the DIART framework, a diarization system based on the online clustering algorithm UIS-RNN-SML, and the end-to-end online diarization system FS-EEND are compared. The lowest latency is achieved for the DIART-pipeline with the embedding model pyannote/embedding and the segmentation model pyannote/segmentation. The FS-EEND system shows a similarly good latency. In general there is currently no published research that compares several online diarization systems in terms of their latency. This makes this work even more relevant.
\end{abstract}


\section{Introduction}
In the machine learning task speaker diarization, the question "who spoke when" is answered for an audio file. This information is an important component of a fully-fledged audio transcription. For this reason, speaker diarization is used in many transcription scenarios such as online meetings, earnings reports, court proceedings, interviews, etc. \cite{park_review_2021}. In some of these scenarios, speaker diarization is required with a low latency. For example, automated stock purchases or sales can be executed immediately based on the transcription of an earnings report \cite{de_castro_mt5b3_2021}. Speaker diarization with a low latency is referred to as online speaker diarization.\\

Most publishers of online speaker diarization systems provide information on the latency of their systems e.g. \cite{coria_overlap-aware_2021} \cite{liang_frame-wise_2023} \cite{xia_turn--diarize_2022}. However, there is still no publication that compares the latency of different online diarization systems on the same hardware with the same test data. This paper aims to fill this gap. In this work, the latency is measured for combinations of different segmentation and embeddings models of the DIART framework \cite{coria_overlap-aware_2021}. In addition, the supervised clustering system UIS-RNN-SML \cite{fini_supervised_2019} and the end-to-end diarization system FS-EEND \cite{liang_frame-wise_2023} are included in the evaluation. This is intended to answer the research question of which online diarization system has the lowest latency from audio input to speaker label output on a uniform hardware. \\
The rest of the paper is structured as follows. In the next section, the basics and related scientific work are presented.  Next the research method, data collection, preprocessing of the data, machine learning models used and the evaluation method are discussed. In the following section, the evaluation results are presented and then discussed. Finally, everything is summarized and possible future work is discussed.

\section{Background and Related Work}
This section lays the scientific foundation for the following sections. It also briefly summarizes some scientific papers that are related to this paper.

\subsection{Speaker Diarization Pipeline}
The classic speaker diarization pipeline usually consists of the three sub tasks: Speech Activity Detection (SAD), Segmentation and Clustering. SAD checks whether the incoming audio segment contains speech. During segmentation, the audio segments are cut so that they only contain one speaker. In clustering, the audio segments are then assigned to known or new speakers \cite{park_review_2021}.

In the past, each sub task was solved by a separate model. By using deep learning methods, several sub tasks or even the entire pipeline can now be automated by a single neural network. Systems that automate the entire pipeline with one model are referred to as end-to-end systems \cite{park_review_2021}.

\subsection{DIART Framework} \label{diart}
In 2021, Coria et al. \cite{coria_overlap-aware_2021} developed an online speaker diarization system that combines an end-to-end approach with a modular approach. The advantage of end-to-end systems is that they can automatically deal with speaker overlaps through supervised training. The disadvantage of end-to-end approaches is that the maximum number of speakers must be known in advance. This is difficult with online speaker diarization, as new speakers can always be added to an infinite input stream. Coria et al. therefore apply the end-to-end approach to individual audio chunks with a limited number of speakers, thereby generating local speaker labels. Subsequently, an incremental clustering algorithm maps the local speaker labels to cross-chunk global speaker labels.

The module for the local speaker labels receives a 5s rolling audio buffer as input. Diarization is performed at intervals of the parameter \textit{step\_size} (e.g. 250ms). The outputs are active speaker propabilities for each frame. These are passed on to the subsequent clustering module.

The clustering module generates a speaker embedding from the active frames for each speaker in the rolling buffer. Constrained incremental clustering then assigns the local speaker embeddings to the global speaker embeddings. If the distance between a local speaker embedding and the existing global speaker embeddings is too great, a new global speaker embedding is created from the local speaker embedding. The global speaker embeddings are then updated with the local speaker embeddings.

\subsection{UIS-RNN-SML}
The online diarization system by Fini et. al. \cite{fini_supervised_2019} is an adapted version of the original UIS-RNN \cite{zhang_fully_2019}. UIS-RNN is a supervised approach for the clustering component of a speaker diarization pipeline. Fini et. al. have developed an adapted loss function and a formula for calculating the parameter \(\alpha\) for the original UIS-RNN algorithm. 
In the original loss function of the UIS-RNN algorithm, the error is calculated using the Mean Square Error (MSE) from the following components:
\begin{itemize}
    \item Average of the outputs of the gated recurrent unit (GRU) instance
    \item Embedding of the next audio sequence
\end{itemize}
In this case, the network only learns a connection between the current observation and the next audio sequence. 
With the adapted loss function, on the other hand, the MSE is composed as follows:
\begin{itemize}
    \item Average of the outputs of the gated recurrent unit (GRU) instance
    \item Average of an unseen collection of embeddings of the current speaker
\end{itemize}

The adapted loss function allows the network to establish a connection between the current observation and the current speaker. This leads to faster convergence, better minima and better generalization. This is also reflected in the measurement results of Fini et. al \cite{fini_supervised_2019}.

The \(\alpha\) parameter of the UIS-RNN algorithm represents the probability that a new speaker will enter the conversation. If the \(\alpha\) is too high, the algorithm overestimates the number of speakers. If \(\alpha\) is too low, the number of speakers is underestimated. In the original UIS-RNN, a constant value is used that is specified by the user. In the adapted version of the UIS-RNN, Fini et. al. have developed a formula that determines \(\alpha\) using the ratio of the count of speaker labels and the count of speaker changes. However, the calculation of \(\alpha\) only leads to a slight improvement in performance \cite{fini_supervised_2019}.

\subsection{FS-EEND}
Frame-wise streaming end-to-end diarization (FS-EEND) is an end-to-end online diarization system introduced by Liang et. al. in 2023 \cite{liang_frame-wise_2023}. The system receives log-mel audio features as input, which consist of one or more audio frames. These are converted into audio embeddings by an embedding encoder. The attractor decoder generates attractors from the audio embeddings.  An attractor is the representation of a speaker. Each speaker embedding is then assigned the corresponding attractor and therefore the corresponding speaker label via a similarity comparison. Both the embedding encoder and the attractor decoder use self-attention in order to achieve better diarization results and to avoid losing information about past frames.

\section{Methodology}
This section provides an overview of the data collection, preprocessing of the data, used machine learning algorithms and the experimental setup.

\subsection{Research Method}
The aim of this work is to analyze which online diarization system has the lowest latency values from audio input to speaker label output on a uniform hardware. Therefore a structured experiment is carried out. For a uniform dataset, the latency is measured for different combinations of segmentation and embedding models of the DIART framework. In addition, the measurement is carried out for the online diarization systems UIS-RNN-SML and FS-EEND. The results are then classified and discussed.

\subsection{Data Collection}
There are several different pretrained segmentation and embedding models for the DIART framework. However, there are no pretrained models for the online diarization systems UIS-RNN-SML and FS-EEND. Corresponding models must therefore be trained for these systems. In this paper the TIMIT dataset \cite{garofolo_john_s_timit_1993} is used for this purpose. For comparable accuracy results as in the original papers, there is not enough training data available in this research work. Therefore, only the latency is reported in this paper. For the accuracy and Diarziation Errors (DER) of the systems, this work refers to the corresponding original paper.

A subset of the Voxconverse testset \cite{chung_spot_2020} is used in this work as a testset for measuring latency. The subset consists of the first four audio files aepyx.wav, aggyz.wav, aiqwk.wav and auzuru.wav. These contain a total of around 20 minutes of audio recordings.

\subsection{Preprocessing}
The DIART framework can directly process the audio files in \textit{wav} format and therefore requires no further preprocessing. The speaker labels can also be provided in \textit{rttm} format to measure the DER.

The implementation of the UIS-RNN-SML, on the other hand, only contains the clustering component and requires audio embeddings for clustering as input \cite{fini_supervised_2019}. Suitable audio embeddings in the form of d-vectors \cite{variani_deep_2014} can be generated with the audio embedding system from Harry Volek \cite{harryvolek_harryvolekpytorch_speaker_verification_2024}.

As described above, FS-EEND is an end-to-end diarization system that can work directly with log-mel audio features. However, the corresponding implementation by Liang et al. \cite{liang_audio-westlakeufs-eend_2024} is currently structured in such a way that it expects the input data in the Kaldi style. Kaldi is a toolkit for speech recognition \cite{povey_kaldi_2011}, which also implements its own format for data preparation. For the conversion of the TIMIT dataset into the Kaldi style, the toolkit provides corresponding scripts that can be used with some adaptations. For the conversion of the test data from the Voxconverse dataset to Kaldi style separate scripts had to be created.

\subsection{Machine Learning Models}
All combinations of the following embedding and segmentation models are tested for the DIART framework. \\
Embedding models:
\begin{itemize}
    \item pyannote/embedding
    \item pyannote/wespeaker-voxceleb-resnet34-LM
    \item speechbrain/spkrec-ecapa-voxceleb
    \item speechbrain/spkrec-xvect-voxceleb
\end{itemize}
Segmentation models:
\begin{itemize}
    \item pyannote/segmentation
    \item pyannote/segmentation@Interspeech2021
    \item pyannote/segmentation-3.0
\end{itemize}
These are the models with the lowest latency in the measurement by Coria et al. \cite{coria_overlap-aware_2021}.

The UIS-RNN-SML implementation \cite{fini_supervised_2019} only contains the online clustering algorithm based on an recurrent neural network (RNN) architecture. For a complete speaker diarization pipeline, the clustering module must be preceded by a SAD and segmentation module. The implementation by Harry Volek \cite{harryvolek_harryvolekpytorch_speaker_verification_2024} is used for this purpose. This implementation is based on a Long Short-Term Memory (LSTM) model for generating d-vectors as audio embeddings. 

The last machine learning algorithm used in this work is the end-to-end diarization system FS-EEND \cite{liang_frame-wise_2023}. The system is based on an encoder-decoder architecture with self-attention modules. 

\section{Experimental Setup}
\subsection{Model Training}
As described above, there are no corresponding pretrained models for the UIS-RNN-SML and FS-EEND systems. Therefore, models are trained with the TIMIT dataset for this work. 

For the UIS-RNN-SML system, both a model for generating the d-vectors and a model the UIS-RNN-SML algorithm must be trained. The implementation by Harry Volek \cite{harryvolek_harryvolekpytorch_speaker_verification_2024} provides corresponding scripts for preprocessing and training with the TIMIT dataset. The training is carried out with the provided default training configuration. The configuration suggests 950 epochs and a learning rate of 0.01 for an LSTM with 768 hidden layers. The resulting model can then be used to create d-vectors for the TIMIT dataset. The d-vectors are then used to train the UIS-RNN-SML. There are also suggested default values for the training of the UIS-RNN-SML. These are 10 epochs, 20,000 train iterations and a learning rate of 0.003 for an RNN with a hidden size of 512.

FS-EEND expects the input data in Kaldi style. The Kaldi toolkit provides scripts to convert the TIMIT dataset into the Kaldi style. Only functions for creating the \textit{reco2dur} file and the \textit{segements} file need to be added here. The training is carried out for 100 epochs, 100,000 warm-up steps and a learning rate of 1 for a Transformer model with 4 encoding and 2 decoding layers. 

The trained models do not have the same quality as the models of the original paper. However, the quality of the models is irrelevant for a pure latency measurement.

\subsection{Evaluation}
To enable a comparable evaluation of the different systems, 250ms chunks are transferred to each system. The evaluation is carried out on an Intel(R) Xeon(R) Gold 5215 CPU @ 2.50GHz. The Python library time is then used to measure the following time span:

\begin{equation}
\begin{split}
latency = & t_{2}(\text{time of the speaker label output}) \\
- & t_{1}(\text{time of the audio chunk input})
\end{split}
\end{equation}

The function \textit{time.perf\_counter} is used for the measurement. The function uses a performance counter with the highest possible resolution on the test system for the time measurement \cite{python_documentation_time_2024}.
\subsubsection{DIART framework}
The DIART framework already offers a way to measure the latency for processing a chunk. Here, only the used time function \textit{time.monotonic} is exchanged for \textit{time.pref\_counter} to ensure comparability with the other systems. In this work, the mean and the standard deviation of the corresponding latency values are reported for each system.

\subsubsection{UIS-RNN-SML}
For UIS-RNN-SML, d-vectors are created from the test audio chunks using the implementation by Harry Volek \cite{harryvolek_harryvolekpytorch_speaker_verification_2024} and then transferred to the UIS-RNN-SML algorithm as input. The current implementation of UIS-RNN-SML expects the test sequence in one piece and generates the corresponding speaker label iteratively for each chunk internally. In order to be able to evaluate UIS-RNN-SML unchanged, the latency measurement must be split up. 
\begin{itemize}
    \item First, the latency for generating the d-vector is measured for each 250ms chunk.
    \item Then the d-vectors are transferred collectively to UIS-RNN-SML.
    \item After that the latency for generating the speaker label is measured per chunk within the UIS-RNN-SML.
\end{itemize}

\subsubsection{FS-EEND}
For the online diarization system FS-EEND, the time span per test step is measured for each 250ms chunk. The latency measurement must also be split up here. The reason for this is that the log-mel features are created in one class and the transformer model is executed in another class. In order to change the code as little as possible, the two time spans are measured separately and then summed up afterwards.

\section{Results}
In this chapter the results of the evaluation of the various systems are presented.
\subsection{DIART Framework}
The evaluation results of the DIART framework models are shown in table \ref{table:1}. The table shows the measurement results for the best 5 model combinations. It can be seen that the lowest latency is achieved with the combination \textit{pyannote/segmentation} and \textit{pyannote/embedding}. The mean latency for this model combination is 0.057065s. It is also noticeable that the embedding models are primarily decisive for the latency.

\begin{table*}[t]
  \centering
    \caption{Evaluation results - DIART framework}
      \begin{tabular}{|>{\columncolor{lightgray!70}}p{1cm}|p{3.4cm}|p{3.4cm}|p{1cm}|p{2.1cm}|p{2.1cm}|}
        \hline
        \rowcolor{lightgray!70} \textbf{rank} & \textbf{segmentation model} & \textbf{embedding model}& \textbf{DER} & \textbf{latency mean in s} & \textbf{latency std in s} \\
        \hline
        1 & pyannote/segmentation &	pyannote/embedding &	44.86 &	0.057065 &	0.003522 \\ 
        \hline
        2 &pyannote/segmentation-3.0	& pyannote/embedding&	48.79&	0.060674	&0.002804 \\
        \hline
        3 &pyannote/segmentation @Interspeech2021	&pyannote/embedding&	50.06&	0.064204&	0.007641 \\
        \hline
        4 &pyannote/segmentation-3.0&	pyannote/wespeaker-voxceleb-resnet34-LM&	48.66	&0.217110	&0.016487 \\
        \hline
        5 &pyannote/segmentation	&pyannote/wespeaker-voxceleb-resnet34-LM&	45.42	&0.218060	&0.010972 \\
        \hline
        6 &pyannote/segmentation @Interspeech2021	&pyannote/wespeaker-voxceleb-resnet34-LM	&49.89	&0.284455	&0.011451 \\
        \hline
        7 &pyannote/segmentation-3.0&	speechbrain/spkrec-ecapa-voxceleb&	49.16&	0.461488	&0.082039 \\
        \hline
      \end{tabular}
  \label{table:1}
\end{table*}

\subsection{UIS-RNN-SML}
The results of the evaluation of the UIS-RNN-SML are shown in table \ref{table:2}. The table shows the measurement results for the first three and the last three chunks as well as the corresponding mean values. Some steps, e.g. 0.25s, are missing. This is due to the fact that the SAD component sorts out all non-speech segments directly \cite{zhang_fully_2019}. The latency for creating the d-vectors is a constant value. However, the latency for clustering a chunk shows a constantly increases proportional to the length of the audio stream. 

\begin{table*}[t]
  \centering
    \caption{Evaluation results - UIS-RNN-SML}
      \begin{tabular}{|p{2.1cm}|p{1.8cm}|p{3.2cm}|p{3.2cm}|p{3.2cm}|}
        \hline
        \rowcolor{lightgray!70} \textbf{time segment} & \textbf{file} & \textbf{latency d-vec creation in s} & \textbf{latency UIS-RNN-SML in s} & \textbf{total latency in s} \\
        \hline
            0.5&	aepyx.wav	&0.325704	&0.010851&0.336555\\
            \hline
            0.75&	aepyx.wav	&0.238017	&0.005043&0.24306\\
            \hline
            1	&aepyx.wav	&0.213074	&0.013827&0.226901\\
            \hline
            ..&	..	&..&	..& ..\\
            \hline           
            1144	&auzru.wav	&0.185197&	9.179082&9.364279\\
            \hline
            1144.25	&auzru.wav&	0.179403&	9.183815&9.363218\\
            \hline
            1144.5	&auzru.wav&	0.180698&	10.07876& 10.259458\\
            \hline
            &&		Mean 0.189569 / Std 0.027847&	& \\
        \hline
      \end{tabular}
  \label{table:2}
\end{table*}

\subsection{FS-EEND}
The results for the latency evaluation of the FS-EEND system can be seen in table \ref{table:3}. The latency per chunk is independent of the total length of the audio file. The mean latency is 0.058293s, which is similar to the best system of the DIART framework.

\begin{table*}[h]
    \centering
    \caption{Evaluation results - FS-EEND}
    \begin{tabular}{|p{2.1cm}|p{1.8cm}|p{3.2cm}|p{3.2cm}|p{3.2cm}|}
    \hline
         \rowcolor{lightgray!70} \textbf{time segment in s} & \textbf{file} & \textbf{log-mel latency in s}& \textbf{FS-EEND latency in s}& \textbf{total latency in s} \\
         \hline
        0	&aepyx.wav &0.002818 &	0.052486 &0.055304 \\\hline
        0.25&	aepyx.wav&0.002239&0.051046	&0.053285\\\hline
        0.5	&aepyx.wav &0.00204&0.046209	&0.04825\\\hline
        ...&	...&	...&...&...\\\hline
        1145&	auzru.wav&0.002147&0.045769	&0.047915\\\hline
        1145.25&	auzru.wav&	0.001864&0.051708&0.053572\\\hline
        1145.5	&auzru.wav&	0.002094&0.049021&0.051115\\\hline
	& &Mean 0.002381 / Std 0.000423&Mean 0.055913 / Std 0.006509&Mean 0.058293 / Std 0.006509\\\hline
    \end{tabular}

    \label{table:3}
\end{table*}

\section{Discussion}
\subsection{DIART Framework}
Table \ref{table:1} shows that pipelines with the embedding model \textit{pyannote/embedding} have the lowest latency. Pipelines with the embedding model \textit{pyannote/wespeaker-voxceleb-resnet34-LM} are in second place and those with the model \textit{speechbrain/spkrec-ecapa-voxceleb} are in third place. The reason for this is the size and depth of the embedding models. The \textit{pyannote/embedding model} is based on a canonical x-vector TDNN-based architecture with approximately 4.3 million parameters \cite{snyder_x-vectors_2018}. \textit{Pyannote/wespeaker-voxceleb-resnet34-LM} is based on ResNet and already has 6.6 million parameters \cite{wang_wespeaker_2023}. \textit{Speechbrain/spkrec-ecapa-voxceleb} has 22.1 million parameters and is based on an ECAPA-TDNN model \cite{desplanques_ecapa-tdnn_2020}. \\

The segmentation models have a much smaller impact on the overall latency, as they are smaller models. The segmentation models are built from a combination of convolutional and LSTM layers with approximately 1.4 million parameters \cite{bredin_end--end_2021}. Since the DIART framework works with a rolling buffer, the latency per chunk remains constant regardless of the length of the audio stream. However, as described in \ref{diart}, incremental clustering is performed for the assignment of local and global speaker lables. Thus, in theory, the latency must increase with the number of speakers. However, this correlation could not be observed with the used test setup.

\subsection{UIS-RNN-SML}
\begin{figure}[ht]
    \centering
    \includegraphics[width=0.9\linewidth]{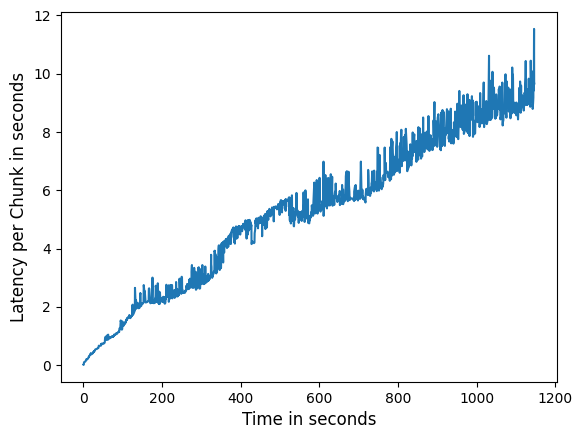}
    \caption{Latency UIS-RNN-SML per chunk}
    \label{fig:latency_uis_rnn_sml}
\end{figure}

\begin{figure}[ht]
    \centering
    \includegraphics[width=0.9\linewidth]{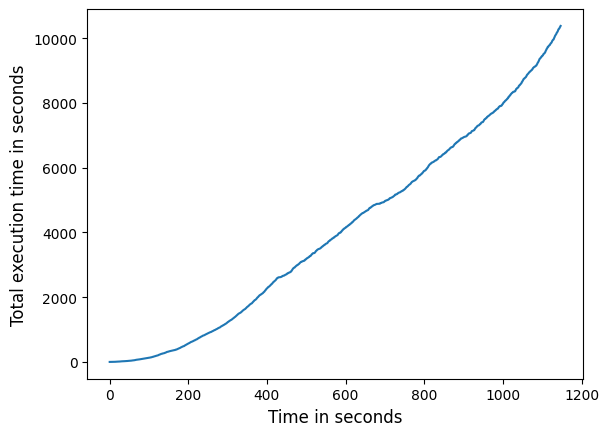}
    \caption{Total execution time UIS-RNN-SML}
    \label{fig:exetime_uis_rnn_sml}
\end{figure}

In the results of the UIS-RNN-SML system, it is noticeable that the latency per chunk increases with the length of the audio stream. The reason for this is that UIS-RNN performs a beam search on the label space for internal decoding. For each new chunk a further beam state is added to the beam search. This increases the decoding time per chunk linearly \(O(t)\) (figure \ref{fig:latency_uis_rnn_sml}). However, the beam search must be performed once for each chunk. This results in a total runtime of \(O(t^2)\) (figure \ref{fig:exetime_uis_rnn_sml}). Due to the increasing latency, UIS-RNN-SML is not suitable for longer audio recordings or audio streams.

\subsection{FS-EEND}
With an average latency of 0.058s, FS-EEND achieved similarly good results in the evaluation as the best pipeline of the DIART framework. The latency remains constant regardless of the length of the audio stream. The self-attention modules of the FS-EEND are used for the correct assignment of already known speakers. Although these modules have a longer memory than for example a LSTM, self-attention modules do not memorize already known speakers forever \cite{liang_frame-wise_2023}. FS-EEND can therefore not be recommended without restriction for very long audio streams. In addition, FS-EEND is a pure end-to-end diarization system. For such systems, the maximum number of speakers must be known in advance in order to produce good results \cite{park_review_2021}. This is only practicable to a limited extent for an online diarization scenario as described in \ref{diart}.

\section{Conclusion}
In this work, various online diarization systems are evaluated on the same hardware with the same dataset in terms of their latency. The DIART framework with the embedding model \textit{pyannote/embedding} and the segmentation model \textit{pyannote/segmentation} proved to be the best system. The end-to-end diarization system FS-EEND has a similarly good latency. For short audio streams, the system with UIS-RNN-SML also has an acceptable latency. However, the latency increases with the length of the audio input. Therefore, UIS-RNN-SML is not recommended for long audio streams. 

In the context of this research work, not enough training data was available to bring the trained models for UIS-RNN-SML and FS-EEND to a comparable performance as the pre-trained models of the DIART framework. Also, FS-EEND is a pure end-to-end system and should therefore have a lower accuracy, if the maximum number of speakers in the inference is higher than the maximum number of speakers while the training. For future work, it would be interesting to train models with a comparable accuracy and then evaluate them in terms of their accuracy-latency-ratio. 

In addition, the architecture of the DIART pipeline suggests that the latency must increase with the number of known speakers. This correlation could not be measured with the evaluation setup of this work. For future work, it would be interesting to test this correlation with a correspondingly large number of speakers. 

In summary, it can be said that several high-performance online speaker diarization systems exist to perform diarization in near real time.

\bibliographystyle{IEEEtran}
\bibliography{references}

\end{document}